\documentclass[10pt,twocolumn,letterpaper,fleqn]{article}

\usepackage[preprint]{cvpr}      

\usepackage{cuted}

\usepackage[dvipsnames]{xcolor}

\definecolor{cvprblue}{rgb}{0.21,0.49,0.74}
\usepackage[pagebackref,breaklinks,colorlinks,citecolor=cvprblue]{hyperref}
\usepackage[capitalize]{cleveref}
\usepackage{graphicx}
\usepackage{algorithm}
\usepackage{float}
\usepackage{stfloats}
\usepackage{mathtools}
\usepackage{algpseudocode}
\usepackage{amsmath}
\DeclareMathOperator*{\argminA}{arg\,min} 
\crefname{section}{Sec.}{Secs.}
\Crefname{section}{Section}{Sections}
\def\paperID{2122} 
\def\confName{CVPR}
\def\confYear{2024}

\title{FED-NeRF: Achieve High 3D Consistency and Temporal Coherence for  \\ \textsl{F}ace Video \textsl{E}diting on \textsl{D}ynamic \textsl{NeRF}}
\vspace{-0.2in}

\author{
  \begin{tabular}[t]{c}
    \textnormal{Hao Zhang} \\
    \textnormal{HKUST} \\
    \textnormal{hzhangcc@connect.ust.hk}
  \end{tabular}
  \quad
  \begin{tabular}[t]{c}
    Yu-Wing Tai \\
    Dartmouth College \\
    yu-wing.tai@dartmouth.edu
  \end{tabular}
  \quad
  \begin{tabular}[t]{c}
    Chi-Keung Tang \\
    HKUST \\
    cktang@cs.ust.hk 
  \end{tabular}
}

\begin{document}
\maketitle

\begin{strip}
    \centering
\includegraphics[width=0.99\linewidth]{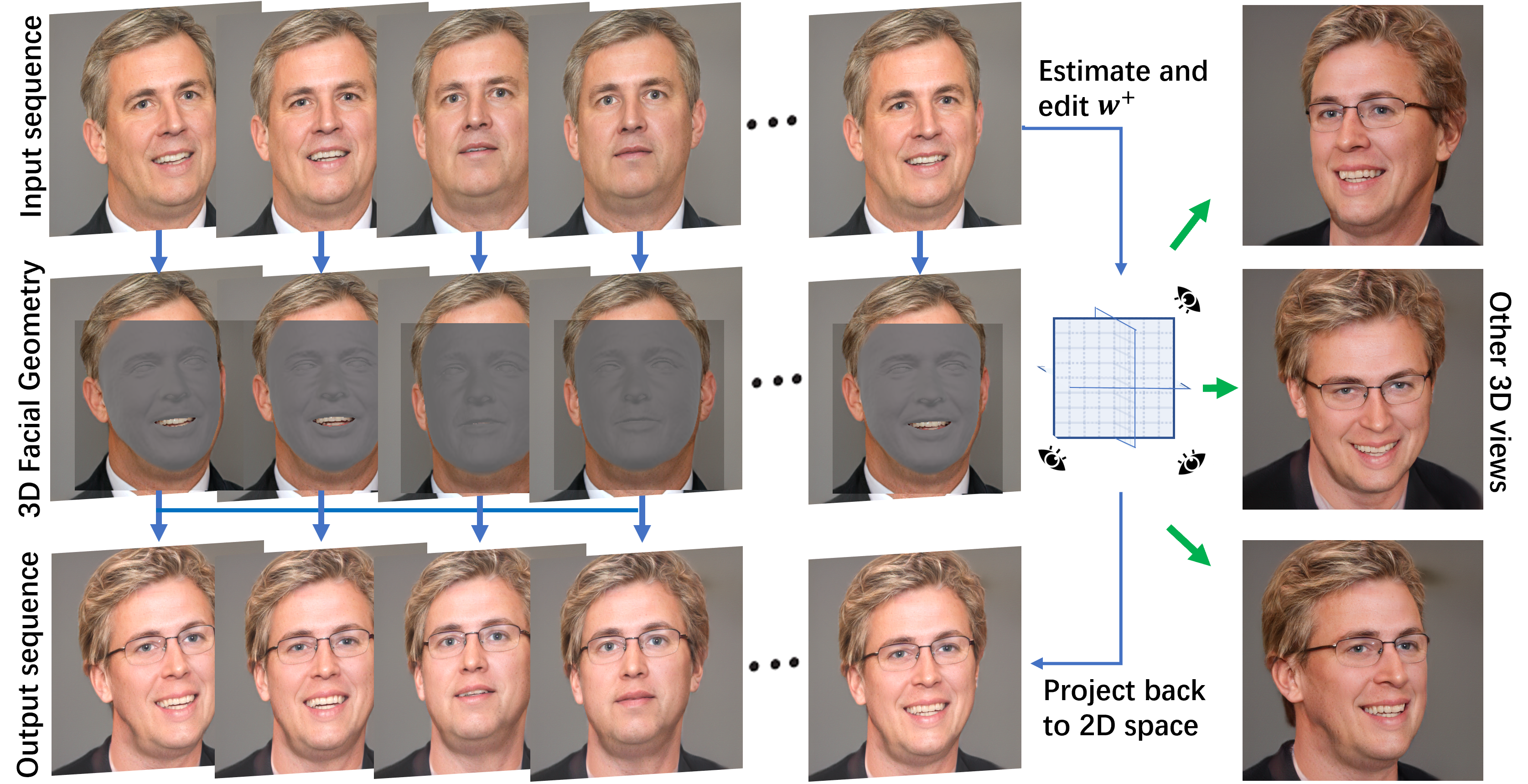}
\vspace{-0.1in}
    \captionof{figure}{{\bf Face video Editing results.} Editing prompts are ``Wear a pair of glasses" and ``Curly hair". Every frame within the output sequence is rendered via the dynamic NeRF, which is precisely controlled by the estimated 3D facial geometry. Furthermore, the other 3D views effectively showcase the consistency of the dynamic NeRF.\vspace{-0.1in}}
    \label{fig:teaser}
\end{strip}

\begin{abstract}
The success of the GAN-NeRF structure has enabled face editing on NeRF to maintain 3D view consistency. However, achieving simultaneously multi-view consistency and temporal coherence while editing video sequences remains a formidable challenge. This paper proposes a novel face video editing architecture built upon the dynamic face GAN-NeRF structure, which effectively utilizes video sequences to restore the latent code and 3D face geometry. By editing the latent code, multi-view consistent editing on the face can be ensured, as validated by multiview stereo reconstruction on the resulting edited images in our dynamic NeRF. As the estimation of face geometries occurs on a frame-by-frame basis, this may introduce a jittering issue. We propose a stabilizer that maintains temporal coherence by preserving smooth changes of face expressions in consecutive frames. Quantitative and qualitative analyses reveal that our method, as the pioneering 4D face video editor, achieves state-of-the-art performance in comparison to existing 2D or 3D-based approaches independently addressing identity and motion. Codes will be released. \end{abstract}    
\section{Introduction}
\label{sec:intro}
Realistic human face synthesis and editing have constituted a prominent research area with their vast range of applications. 
Previous research employed subspace deformation or face morphing techniques~\cite{Thies_2016_CVPR,yang2012face, dale2011video} to achieve expression transfer and reenactment with 
impressive results. However, these methods were limited to blending existing faces and were unable to add substantially new or alter facial features realistically while preserving identity.  
With the debut of
Generative Adversarial Networks (GANs)~\cite{GANs},  
the latent space of GANs, which possesses desirable properties such as perceptual path length and linear separability as elaborated in~\cite{StyleGAN}, have been employed to make face editing more flexible and versatile. Several recent studies, including ~\cite{Patashnik_2021_ICCV,wu2021stylespace,song2022diffusion}, have demonstrated how to edit a face image in  GAN's latent space. 

When editing a face in a {\em video}, possibly captured in the wild, it is crucial to ensure temporal coherence and 3D view consistency. To achieve temporal coherence, researchers in~\cite{alaluf2022times,10012505,tzaban2022stitch,yao2021latent,xu2022videoeditgan} have extended image editing to video editing by adding constraints between consecutive frames. Meanwhile, to ensure 3D view consistency while editing face features, researchers in~\cite{sun2022ide,sun2022fenerf,zhang2023fdnerf} have leveraged the GAN-NeRF structure~\cite{Schwarz2020NEURIPS, GIRAFFE,piGAN2021}. However, these methods are unable to simultaneously and robustly guarantee 3D view consistency and temporal coherence.

Thus it is quite imperative to elevate the editing process to a 4D space to achieve spatio-temporal coherence. Dynamic NeRF encompasses two desirable mechanisms in this regard: the first consists of a canonical space and a deformation space as outlined in various studies~\cite{pumarola2020d,park2021hypernerf,park2021nerfies,NEURIPS2022_cb78e6b5}, while the second involves conditioning the original neural radiance fields on time-related variables~\cite{wang2021neural, xian2021spacetime,zhuang2022mofanerf}. To attain better disentanglement of shape and motion, the first mechanism is adopted into our dynamic NeRF representation. In order to leverage the inherent editability within the latent space, the GAN-NeRF model incorporating the first mechanism~\cite{yue2022anifacegan, Xu_2023_CVPR_OmniAvatar} emerges as a better option. Notably, the study by~\cite{Xu_2023_CVPR_OmniAvatar} uses the FLAME model~\cite{FLAME:SiggraphAsia2017} to represent the geometry of the deformation field, which offers higher expressiveness compared to 3DMM~\cite{3DMM_SIGGRAPH99} used in~\cite{yue2022anifacegan}.

The studies in~\cite{yue2022anifacegan, Xu_2023_CVPR_OmniAvatar} proposed a structure for animating a talking head using a given latent code and a sequence of continuous expression codes. However, in video cases, obtaining the ground truth latent and expression codes remains challenging. Although GAN-inversion methods can generate the latent code given the estimated expression codes provided by off-the-shelf expression estimators ~\cite{EMOCA:CVPR:2021,DECA:Siggraph2021, tran2016regressing}, the resulting edited video may not achieve satisfactory performance due to inaccuracies accumulated during the estimation process. 

To address editable dynamic face NeRF with the above issues, we introduce FED-NeRF, a novel face video editing architecture that thoroughly utilizes the information embedded in video sequences to restore the latent code of GAN-NeRF space, and the sequences of expression codes for each frame as well. To accurately restore the latent code, we first extract $w^+$ features using an encoder based on~\cite{yuan2023make} as the backbone for different frames. Next, we apply a cross-attention layer to these $w^+$ features to aggregate them into a single $w^+$ output. To predict sequences of expression codes, we modify the FLAME encoder of EMOCA~\cite{EMOCA:CVPR:2021} and incorporate it into the Omniavatar backbone~\cite{Xu_2023_CVPR_OmniAvatar} as the FLAME estimator during the training process. Since FLAME controls are estimated on a frame-by-frame basis, we introduce an algorithm that leverages the differentiability of the Catmull–Rom spline to stabilize the sequence of FLAME controls. Together with the edited $w^+$ by our modified Latent mapper, the edited video sequences can be produced. In summary, our main technical contributions are:
\begin{itemize}
\item We propose a latent code estimator that utilizes multi-frames as input and predicts accurate $w^+$ values that are applicable across a wide range of 3D views and face expressions.

\item We propose a 3D face geometry estimator that accurately extracts face shape, expressions, and neck rotations from video sequences.

\item We propose an algorithm that can effectively stabilize the transition of face geometry between consecutive frames.

\item We modify the Latent mapper introduced by StyleClip ~\cite{Patashnik_2021_ICCV} to enable its seamless integration with the Omniavatar backbone.~\cite{Xu_2023_CVPR_OmniAvatar}.

\end{itemize}
Consequently, 
with our new technical contributions, casual users can easily edit facial features and expressions within a large range using simple prompts, while preserving the face's identity and the rest of the given video, producing natural video results using the proposed editable 4D face NeRF. Moreover, as 3D consistency is naturally guaranteed, where the edited images can be immediately in used multiview stereo for 3D face reconstruction. See Figure~\ref{fig:teaser}.



\section{Related Work}
\label{sec:formatting}


\noindent\textbf{Video Editing in 2D space} \quad
Generative Adversarial Networks (GANs)~\cite{GANs} contribute to arguably the first breakthrough in contemporary 2D image generative methods, among which StyleGAN~\cite{StyleGAN} and its variants~\cite{StyleGAN3, StyleGAN2} stand out due to their expressive and well-disentangled latent spaces. Editing a single image via the latent space has been analyzed and shown to be successful in~\cite{Patashnik_2021_ICCV} and~\cite{wu2021stylespace}. The straightforward approach~\cite{tzaban2022stitch} for editing a video is frame-by-frame processing in the same editing direction in the latent space. However, the same editing step cannot guarantee coherence among frames across all the given features, especially for high-frequency facial textures such as beard, mustache, hair, etc. To enhance the temporal coherence and avoid shape distortions between frames, in~\cite{10012505} the StyleGAN2~\cite{StyleGAN2} latent vectors of human face video frames are disentangled to decouple the appearance, shape, expression, and motion from identity. In~\cite{xu2023rigid} learning a temporally compensated latent code was proposed, which found  incoherent noises lie in the high-frequency domain can be disentangled from the latent space. To remove the inconsistency after attribute manipulation, an in-between frame composition constraint was adopted. In addition to GAN models, diffusion models have also be employed for face editing in video sequences. In~\cite{Kim_2023_CVPR} the authors proposed a video editing framework based on diffusion autoencoders, which can effectively decompose identity and motion features from a given video. Nonetheless, the fundamental limitation of 2D video editing lies in its  disregard of 3D geometry information during the editing process. This neglect results in shape distortion and feature alteration in side views, as depicted in Fig.~\ref{fig:comparison1}. Furthermore, achieving multi-view consistency cannot be achieved, as illustrated in Fig.~\ref{fig:comparison2} and Tab.~\ref{tab:comparison_table}.

\noindent\textbf{Video Editing in NeRF space} \quad The Neural Radiance Field (NeRF)~\cite{mildenhall2020nerf}, an implicit neural representation, has become the predominant approach in 3D generation. This method offers several advantages, including continuity, differentiability, compactness, and exceptional novel-view synthesis quality,  distinguishing itself  from conventional, explicit and discrete mesh and point cloud techniques.

GRAF~\cite{GRAF} combines implicit neural rendering with GAN to create a generalizable NeRF. PiGAN~\cite{pi-GAN} employs SiREN~\cite{SiREN} to condition the implicit neural radiance field on the latent space. While 3D consistency is assured, volumetric rendering necessitates substantial computation. With limited computation, the image quality of these methods remains inferior to that of state-of-the-art 2D GANs. Consequently, numerous recent approaches adopt hybrid structures. StyleNeRF~\cite{StyleNeRF} applies volume rendering in the early feature maps in low resolution, followed by upsampling blocks to generate high-resolution images. In contrast to employing volume rendering in early layers, EG3D~\cite{EG3D} performs the operation on a relatively high-resolution feature map using a hybrid representation for 3D features generated by StyleGAN2~\cite{StyleGAN2} backbone, named tri-plane, which can incorporate more information than an explicit structure such as voxel. Given these advancements, 3D human face reconstruction achieves multi-view consistency and high-quality 3D generation. 

Utilizing high-quality GAN-NeRF generation models,  editing  3D facial structures has become increasingly feasible and promising. In~\cite{sun2022ide,sun2022fenerf} an interactive approach was adopted for editing 3D faces, allowing users to draw directly on 2D images. In~\cite{zhang2023fdnerf} a method was introduced on diffusion models for semantically editing facial NeRFs based on a target text prompt.  
Although editing faces on NeRF ensures multi-view 3D consistency, temporal or 4D coherence remains an issue.



\begin{figure}
    \centering
    \includegraphics[width=\linewidth,height=4.5cm]{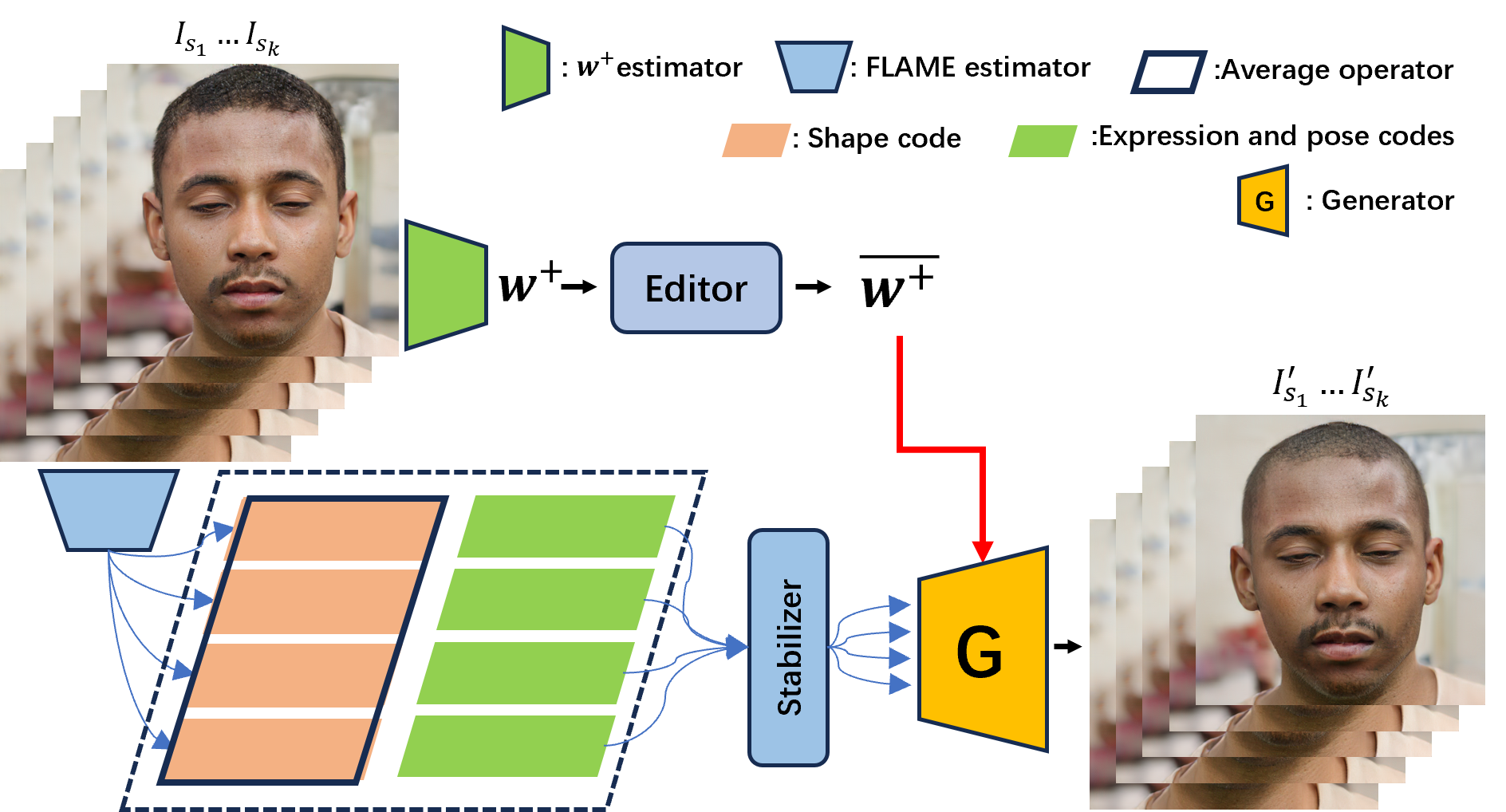}
    \vspace{-0.1in}
    \caption{{\bf The Overview of our model.} Given a video sequence, Our model will estimate a latent code $w^+$ and FLAME controls. The editor will subsequently modify the $w^+$ as $\bar{w^+}$ in accordance with a given text prompt. The Stabilizer then ensures the temporal consistency of the FLAME controls. Finally, the edited video sequence can be produced under the guidance of the stabilized FLAME controls and $\bar{w^+}$ }
    \label{fig:pipeline}
    \vspace{-0.2in}
\end{figure}

\section{Method}
Figure~\ref{fig:pipeline} shows the overall framework. Given an input video, the Latent Code Estimator encodes the detailed face information embedded in the multiple frames into the latent code $w^+$ (Sec.~\ref{sec:Latent_Code_Estimator}). The Face Geometry Estimator extracts the shapes, expressions, and rotations of the jaw and neck from each frame (Sec~\ref{sec:Face Geometry Estimator}). Since the face geometry is estimated individually on each frame, the Stabilizer is proposed to ensure coherence across frames (Sec.~\ref{sec:stabilizer}). In order to perform semantic editing of the facial features, we modify the Latent Code $w^+$ using the Semantic Editor (Sec.~\ref{sec:Semantic Editor}). Subsequently, with the integration of coherent facial geometries and a refined latent code, a photo-realistic edited video of exceptional fidelity can be produced.  We use the Omniavatar~\cite{Xu_2023_CVPR_OmniAvatar}, a dynamic GAN-NeRF structure, as our Generator (Sec.~\ref{sec:Preliminaries}). Our training and test data sets are described in Sec.~\ref{sec:training_data}.

\subsection{Preliminaries} 
\label{sec:Preliminaries}
The Omniavatar~\cite{Xu_2023_CVPR_OmniAvatar} utilizes a 3D-aware generator, EG3D~\cite{EG3D}, as the canonical space representation to achieve photo-realistic and multiview consistent image synthesis. Notably, Omniavatar can disentangle control of face geometric attributes from image generation by employing a 3D statistical head model, FLAME~\cite{FLAME:SiggraphAsia2017}. Essentially, the pertinent deformation from the canonical space to the desired shapes and expressions is encapsulated by a trained deformable semantic SDF around the FLAME geometry. Specifically, a photo-realistic human face image $I_{RGB} (w^+|c, p)$ can be generated by a given latent code $w^+$, a camera pose $c$, and a FLAME control $p = (\alpha, \beta, \theta)$, which consists of shape code $\alpha$, expression code $\beta$, jaw and neck pose $\theta$.


\subsection{Training data}
\label{sec:training_data}

Our objective is to reconstruct the $w^+$ and facial geometry from a video sequence. The majority of existing talking face datasets do not provide ground-truth facial geometry for each frame within the sequences. Thus, we utilize the Omniavatar~\cite{Xu_2023_CVPR_OmniAvatar} to synthesize multi-view images with various expressions for each subject. By randomly sampling $n$ points from the Gaussian distribution and then transforming them through the mapping function, we obtain $n$ latent codes $w_i^+, i \in [0,n-1]$. For each $w_i^+$, 60 FLAME controls $P_i^0,.., P_i^{59}$ are randomly sampled from a large collection of 3D deformed FLAME datasets. Our training dataset $\mathbf{D}$ is thus obtained, where 
we sample $n=30,000$ as the training set, and $n=300$ as the test set. As the FLAME control includes head rotation, we set all camera poses correspond to the frontal view: 
\begin{eqnarray*}
    \label{eq:DataSet}
    \mathbf{D} = \lbrace (w_0^+, I_{RGB} (w_0^+|c, p_0^0), \cdots, I_{RGB} (w_0^+|c, p_0^{59})),\cdots, \\
     ( w_{n-1}^+, I_{RGB} (w_{n-1}^+|c,p_{n-1}^0), \cdots, I_{RGB} (w_{n-1}^+|c, p_{n-1}^{59})) \rbrace
\end{eqnarray*}

\begin{figure*}[t]
    \centering
    \includegraphics[width=\linewidth]{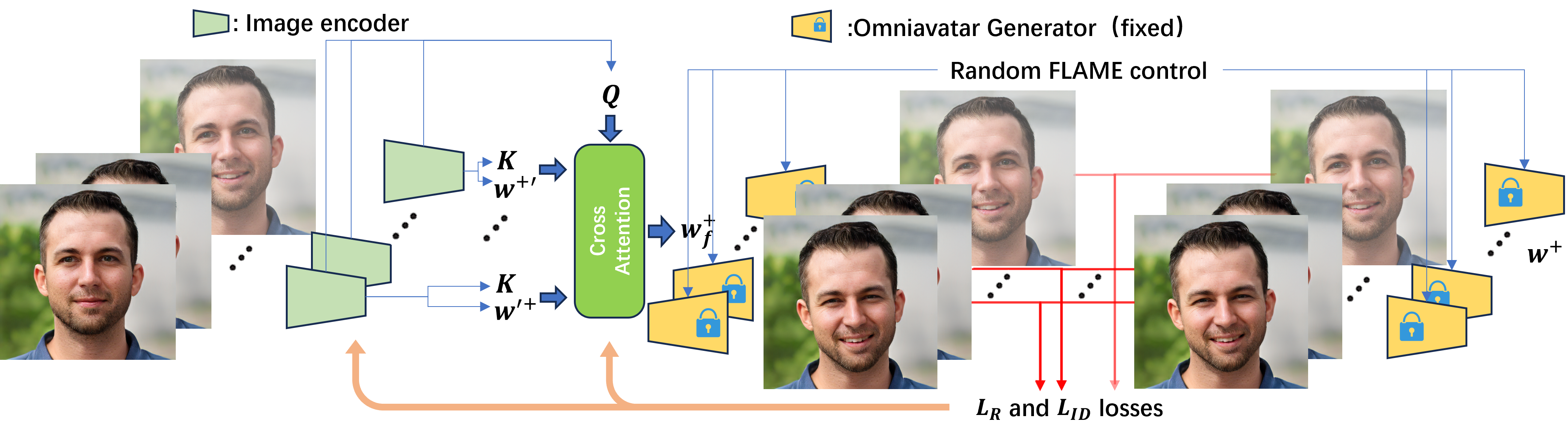}
    \vspace{-0.2in}
    \caption{{\bf The structure of Latent Code Estimator.} Given a video sequence, the Image encoder will extract features for each individual frame, which are then aggregated via the Cross Attention layer to produce a singular latent code output denoted as $w_f^+$. The Losses $\mathcal{L}_R$ and $\mathcal{L}_{\mathcal{ID}}$ are computed across multiple pairs of rendered images by utilizing the estimated $w_f^+$ and ground truth $w^+$ respectively. }
    \label{fig:latent_code_estimator}
    \vspace{-0.1in}
\end{figure*}

\subsection{Latent Code Estimator}
\label{sec:Latent_Code_Estimator}
To aggregate the identity information from the video sequence, we propose the Latent Code Estimator shown in Figure~\ref{fig:latent_code_estimator}. Inspired by~\cite{yuan2023make}, for each frame, we extract the tuple $(Q, K, w^{+\prime})$ according to the backbone’s pyramid features progressively. Here, we also choose Swin-transformer~\cite{liu2021Swin} as the backbone, and further add attention modules at different scale feature layers for different level latent codes, which are concatenated together as the $w^+$ output. The $Q, K$ are extracted from the last layer of the pyramid features by MLP layers, since the $Q, K$ contain the information of how to merge multiple $w^{+\prime}$'s, which is high-level information and thus should be extracted from the latter layers of the pyramid features. After obtaining the tuples $(Q_i, K_i, w_i^{+\prime}),i\in [0,m]$, $m$ is the number of the input frames, the $Q_i, i \in [0,m]$ are averaged to $\overline{Q}$, a Multi-Head Cross Attention layer is applied on the tuples to get the final $w_f^+$
\begin{align*}
     w_f^+ = \mbox{MultiHead}(\overline{Q}, \mathbf{K}, \mathbf{w^+}) \\
     \mbox{where~}  \mathbf{K} = [k_0,..,k_m]^{\intercal}, \mathbf{w^+} = [w_0^{+\prime},..,w_m^{+\prime}]^{\intercal}
\end{align*}

To fully disentangle the $w^+$ estimation with the facial geometry sample, $t$ FLAME controls $p_0,..., p_{t-1}$ are randomly sampled from the large collection of 3D deformed FLAME datasets, when calculating the following Loss function. As shown in Figure~\ref{fig:latent_code_estimator}, the Reconstruction Loss $\mathcal{L}_R$ and ID Loss $\mathcal{L}_{\mathit{ID}}$ are used to measure the differences between the rendered images using $w_f^+$ and the rendered images using ground truth $w^+$:
\begin{eqnarray}
    \mathcal{L}_R = \sum_i\left(||V(I_{RGB}(w^+,p_i))- V(I_{RGB}(w_f^+,p_i))||_2^2\right)\\
    \label{eq: Reconstruction Loss}
    \mathcal{L}_{\mathit{ID}} = \sum_i\left(1 - \langle R(I_{RGB}(w^+,p_i)),R(I_{RGB}(w_f^+,p_i))\rangle\right)
    \label{eq: IDLoss}
\end{eqnarray}
where $V(\cdot)~$ is VGG16 image encoder~\cite{simonyan2015deep}, $R$ is the pretrained ArcFace network~\cite{Deng_2019_CVPR}. The total loss function is thus $\mathcal{L} = \mathcal{L}_R + \mathcal{L}_{\mathit{ID}}$. During training, the Omniavatar Generator $I_{RGB}$ is fixed and the image encoder and Cross Attention layer will be updated. Since the camera poses $c$ remain fixed towards the frontal view, it is omitted from the above equations.

\subsection{Face Geometry Estimator}                                                        
\label{sec:Face Geometry Estimator}
\begin{figure}
    \centering
    \includegraphics[width=\linewidth,height=5.0cm]{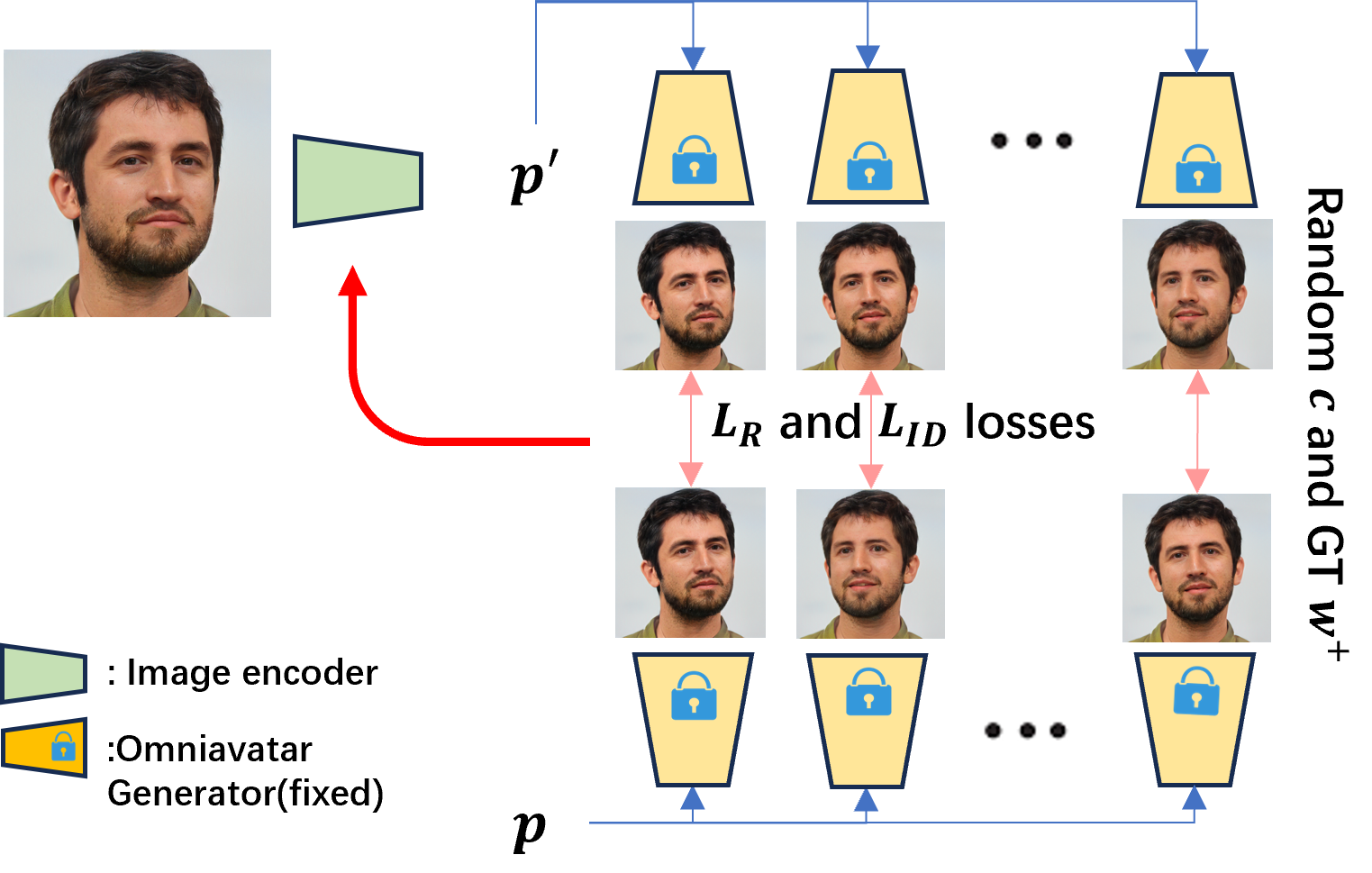}
    \vspace{-0.3in}
    \caption{{\bf The structure of the Face Geometry Estimator.} An estimation of the FLAME control $p^{\prime}$ can be obtained from an input image. Subsequently, pairs of images can be rendered with randomly sampled camera poses, and these losses can be computed based on these pairs.}
    \label{fig:face geometry estimator}
    \vspace{-0.2in}
\end{figure}

Since current facial geometry predictors~\cite{EMOCA:CVPR:2021,DECA:Siggraph2021} are not designed to deform the canonical space of a given dynamic NeRF, the results of directly applying these methods are distorted. To tackle this problem, we propose the framework depicted in Figure~\ref{fig:face geometry estimator}: the image encoder is modified based on~\cite{EMOCA:CVPR:2021}, which first factorizes the input images into facial geometry (represented by FLAME controls), albedo, lighting, additional expression codes, etc. Given these factors, one can differentiably render an output image that should look similar to the input. As albedo, lighting, and detailed face texture are embedded in the latent code $w^+$, only the attributes related to predicting facial geometry are kept while the rest of the EMOCA are discarded. After obtaining the FLAME controls, together with the latent code $w^+$, an output image can be rendered by the Omniavatar Generator. To disentangle the facial geometry with the view direction of the input image, random camera poses $c_i$ are used when rendering the output images. Here, $w^+, p^{\prime}$, and $ p$ denote the ground truth latent code, the estimated FLAME control, and the ground truth FLAME control.
\begin{eqnarray*}
    \mathcal{L}_R = \sum_i\left(||V(I_{RGB}(w^+|c_i,p^{\prime}))- V(I_{RGB}(w^+|c_i,p^{\prime}))||_2^2\right)\\
    \label{eq: Reconstruction Loss}
    \mathcal{L}_{\mathit{ID}} = \sum_i\left(1 - \langle R(I_{RGB}(w^+|c_i,p^{\prime})),R(I_{RGB}(w^+|c_i,p^{\prime}))\rangle\right)
    \label{eq: IDLoss}
\end{eqnarray*}
Similarly, the total loss function is $\mathcal{L} = \mathcal{L}_R + \mathcal{L}_{\mathit{ID}}$. During the training process, the Face Geometry Estimator will be updated, and the Omniavatar Generator $I_{RGB}$ is fixed.

\subsection{Stabilizer}
\label{sec:stabilizer}
Facial geometries are predicted individually on each frame in Sec.~\ref{sec:Face Geometry Estimator}. Thus continuity among consecutive frames cannot be guaranteed. As human motions are largely continuous (second-order differentiable by Newton's Law), we propose to use the Catmull–Rom spline to enforce smooth motion across the video sequence.

Denote  $\mathcal{P}= [p_0,p_1,\cdots,p_{n-1}]^{\intercal}$ as the FLAME controls estimated by Sec.~\ref{sec:Latent_Code_Estimator} at the $n$ timestamps in a video sequences. We aim to estimate a smooth motion sequence $\mathcal{\hat{P}}= [\hat{p}_0,\hat{p}_1,\cdots,\hat{p}_{n-1}]^{\intercal}$ from $\mathcal{P}$. Note that $p_i, \hat{p_i} \in \mathbb{R}^{106}$ (100 dimensions for expression control and 6 dimensions for rotation control) and $\mathcal{P},\mathcal{\hat{P}} \in \mathbb{R}^{n \times 106}$.  We split the sequence $\mathcal{P}$ into $m$ sub-sequences $\mathcal{P}_0, \cdots \mathcal{P}_{m-1}$ with the length $\lfloor n/m \rfloor$ by selecting one FLAME control from every $m$ consecutive FLAME controls. $\mathcal{P}_0, \cdots \mathcal{P}_{m-1} \in \mathbb{R}^{\lfloor n/m \rfloor \times 106}$:
\begin{flalign}
    &\mathcal{P}_0 = [\hat{p}_0,\hat{p}_{m},\hat{p}_{2m},\cdots,\hat{p}_{n-m}] \nonumber \\ 
    &\mathcal{P}_1 = [\hat{p}_1,\hat{p}_{m+1},\hat{p}_{2m+1},\cdots, \hat{p}_{n-m+1}] \nonumber \\
    & ... \nonumber \\
    &\mathcal{P}_{m-1} = [\hat{p}_{m-1},\hat{p}_{2m-1},\hat{p}_{3m-1},\cdots, \hat{p}_{n-1}]
    \label{eqn:seq}
\end{flalign}   

We use the Catmull–Rom spline to form $m$ functions $f_0^i, f_1^i, \cdots, f_{m-1}^i$  using the above $m$ subsequences respectively for the dimension $i\in [0,106 -1]$. Thus, for dimension $i$,  we obtain $m$ estimations from the corresponding $m$ functions at every timestamp. We calculate the distance $D_j^i(t)$ between an estimation $f_j^i(t), j \in [0,m-1]$ with the rest of the estimations at timestamp $t \in [0, n-1]$. 
\begin{equation}
    D_j^i(t) = \sum_{j^\prime =0,\cdots,m-1} \left|f_j^i(t) - f_{j^\prime}^i(t)\right|
 \label{eqn:distance}
\end{equation}  

The function $ f_j^i$ whose distance is ranked within the top two-thirds of  $D_j^i(t), j \in [0,m-1]$ in ascending order will be averaged as the value of $\hat{p}_t^i$ which denotes the value of the $i$th dimension of $\hat{p}_t$. The last one-third is regarded as outliers and will be discarded. After calculating $\hat{p}_t^i$ for all of the dimensions and all of the timestamps,  $\mathcal{\hat{P}}$ is computed.  Algorithm~\ref{alg:Stabilizer} provides the detailed procedure. The hyperparameter $m$ governs the degree of smoothness in the model. An increased value of $m$ yields a more seamless transition, albeit at the expense of reduced fidelity, and vice versa. (Please note that the $argmin$ operator  will return a list of the indices in ascending order based on the value of $D_j^i(t)$.)

\begin{algorithm}
\caption{Stabilizer}\label{alg:Stabilizer}
\begin{algorithmic}
\Require $\mathcal{P}$ and a hyperparameter $m$
\State Initialize $\mathcal{\hat{P}}$
\State Split $\mathcal{P}$ into $\mathcal{P}_0,.., \mathcal{P}_{m-1}$ according to (\ref{eqn:seq}).
\For{i = 0,..,106-1}
    \State Form $f_0^i, f_1^i, \cdots , f_{m-1}^i$ by interpolating \State $\mathcal{P}_0^i,\cdots, \mathcal{P}_{m-1}^i$ using Catmull–Rom spline.
    \For{$t = 0, \cdots,n-1$}
        \For{$j = 0,\cdots,m-1$}
            \State Compute $D_j^i(t)$ according to (\ref{eqn:distance}).
        \EndFor
        \State $l = \argminA_{j \in [0,\cdots,m-a]} D_j^i(t)$. 
        \State $l = l\left[:\lfloor {\frac{2m}{3}} \rfloor\right]$
        \State $\hat{p}_t^i = {\frac{1}{\lfloor {\frac{2m}{3}} \rfloor}} \sum_{v =l[0],\cdots,l[-1]} f_{v}^i (t)$
    \EndFor
\EndFor
\end{algorithmic}
\end{algorithm}


\subsection{Semantic Editor}
\label{sec:Semantic Editor}
We perform semantic editing in the $w^+$ space, where several off-the-shelf methods already exist~\cite{Patashnik_2021_ICCV,wu2021stylespace,zhang2023fdnerf,sun2022fenerf,shen2020interfacegan}. In this study, we employ the Latent Mapper introduced in StyleClip~\cite{Patashnik_2021_ICCV}, as it offers a short inference time of $75ms$ when pre-trained for a particular text prompt. The backbone of the StyleClip is the 2D StyleGAN~\cite{Karras2019stylegan2}. In our work,
the StyleGAN backbone is replaced by the Omiavatar Generator. Since all expressions are deformed with respect to 
the canonical space, which means once the canonical space is edited, all expressions of this person will be edited accordingly. Thanks to this property, our editing is 3D-view consistent and temporally coherent. Thus, the facial geometry control $p$ can be set as zero which corresponds to the canonical space. The latent code $w^+$ is split into three groups (coarse, medium, and fine), or $w^+ = (w_c, w_m,w_f)$. We adopt the same structure of the Latent Mapper as StyleCLIP.  $\mathcal{L}_{\mathit{CLIP}}$ is modified as the following:
\begin{eqnarray}
    \hat{w}^+ = w^+ + M_t(w^+) \\
 \label{eqn:mapper}
    \mathcal{L}_{\mathit{CLIP}} = \mathcal{D}_{\mathit{CLIP}}(I_{RGB}(\hat{w}^+|c_0,p_0), t)
 \label{eqn:L_{clip}}
\end{eqnarray}  
where the camera pose $c_0$ is towards the frontal face, $p_0$ is $\mathit{0}$, $M_t(\cdot)$ is the Latent Mapper, and $\mathcal{D}_{\mathit{CLIP}}(\cdot,\cdot)$ is the cosine distance between the CLIP embeddings of the input image and input text prompt ~\cite{radford2021learning}. The $\mathcal{L}_{\mathit{ID}}$ is modified as the following:
\begin{equation}
    \mathcal{L}_{\mathit{ID}} = 1 - \langle R(I_{RGB}(\hat{w}^+ |c_0,p_0)),R(I_{RGB}(w^+,|c_0,p_0))\rangle
\end{equation}

The total Loss function is analogous to that utilized in StyleCLIP. For a comprehensive elucidation, please refer to the supplementary materials.

\begin{figure*}[t]
\vspace{-0.5in}
    \centering
    \includegraphics[width=0.90\linewidth]{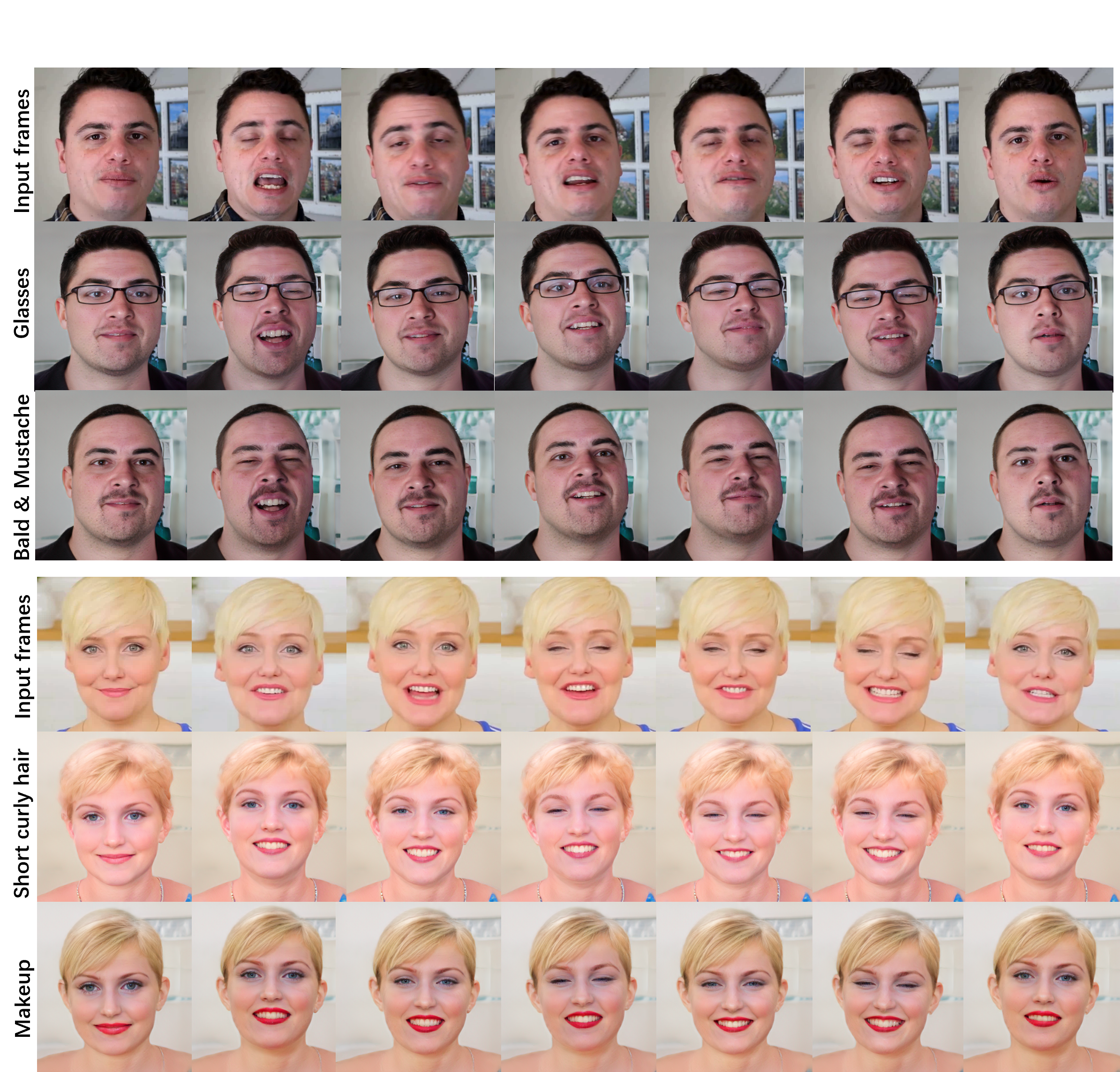}
    \vspace{-0.1in}
    \caption{{\bf More in-the-wild editing results.} These  examples show 
    that our model can achieve 3D consistency even when performing certain edits that alter the facial geometry, such as ``Wear a pair of glasses", ``Short curly hair", and so on.}
    \label{fig:Results2}
    \vspace{-0.1in}
\end{figure*}
\section{Experiments}
\subsection{Editing in-the-wild video sequences}
Our method is capable of achieving good editing results for real-world cases, as demonstrated in Figure~\ref{fig:Results2}. This is due to the fact that Omniavatar ~\cite{Xu_2023_CVPR_OmniAvatar} is trained on FFHQ ~\cite{karras2019style}, a human face dataset containing 70,000 in-the-wild images. The synthesized images generated by Omniavatar exhibit a distribution that is similar to that of real-world cases. Our approach provides greater flexibility in video editing compared to other methods~\cite{tzaban2022stitch,xu2022videoeditgan, Kim_2023_CVPR, yao2021latent} due to the complete disentanglement between facial geometry and face semantic features. This allows us to explicitly edit facial expressions in a video by modifying the FLAME controls, whereas other methods are limited to semantic editing. Additional examples can be found in the supplementary materials.

\begin{figure*}[t]
\vspace{-0.2in}
    \centering
    \includegraphics[width=0.90\linewidth]{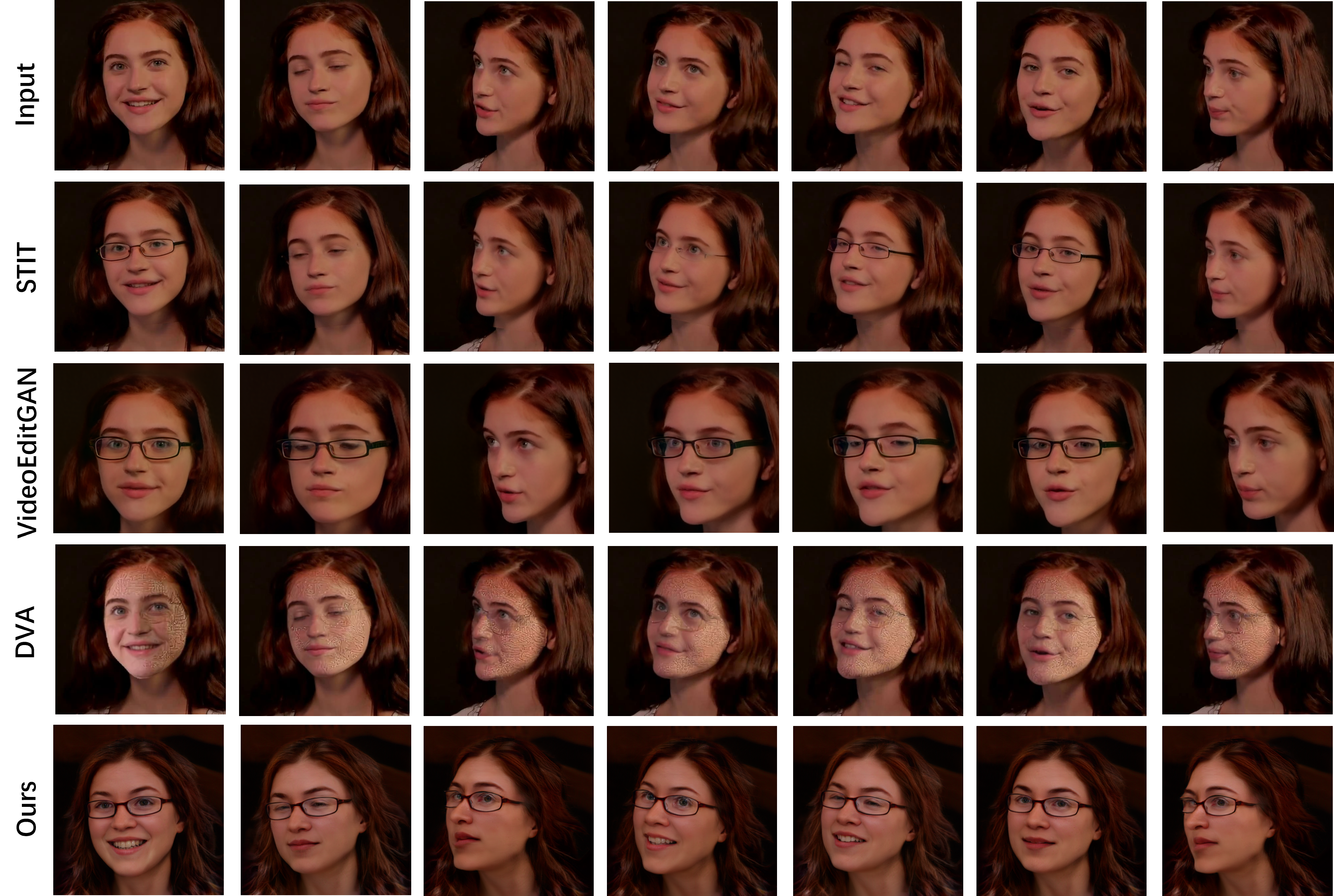}
    \vspace{-0.1in}
    \caption{{\bf Qualitative comparison.} The editing prompt is ``wear a pair of glasses". Note that our method achieves the highest level of 3D view consistency on the edited feature compared to other state-of-the-art methods. STIT, VideoEditGAN, and DVA are the abbreviation for ~\cite{tzaban2022stitch}, \cite{xu2022videoeditgan}, and \cite{Kim_2023_CVPR}}
    \label{fig:comparison1}
    \vspace{-0.1in}
\end{figure*}

\subsection{Comparison}
We conducted a comparative analysis between our proposed method and other existing techniques that aim to facilitate video editing. For ease of reference, we abbreviated the works~\cite{tzaban2022stitch}, \cite{xu2022videoeditgan}, and \cite{Kim_2023_CVPR} as STIT, VideoEditGAN, and DVA, respectively. To evaluate the performance of each method, we randomly selected a talking head video sequence from YouTube. As depicted in Fig.~\ref{fig:comparison1}, the individual in the input sequence turns her face to the side, posing a challenge for all methods to generate a consistent editing outcome in response to the prompt ``Wear a pair of glasses". Notably, both STIT and VideoEditGAN, being 2D GAN-based methods that operate solely in the 2D GAN space and lack awareness of 3D information, fail to ensure the 3D consistency of the human face. As a result, they both produce frames in which the subject is not wearing glasses, or the eyeglasses deform unnaturally 
across different frames. In contrast, our proposed method exhibits superior performance in this scenario. To further investigate, we conduct Sec.~\ref{sec:3D view consistency} and Sec.~\ref{sec:Temporal Coherence} with respect to the 3D view consistency and temporal coherence. In addition to our superior performance, our method also boasts a significantly shorter inference time compared to others. This is due to our utilization of an encoder to locate the latent space, which allows for processing a video in just the time of one forward pass. In contrast, other methods require iterative algorithms such as GAN-inversion for STIT and VideoEditGAN, or diffusion iteration ~\cite{rombach2021highresolution} for DVA. For a video sequence containing 100 frames, our processing time is approximately 3 minutes, whereas STIT and VideoEditGAN require around 3 hours, and DVA requires around 4 hours. All experiments were conducted using an RTX3090.

\subsubsection{3D  View Consistency}
\label{sec:3D view consistency}
\begin{figure}[h]
    \centering
    \includegraphics[width=\linewidth,height=6.0cm]{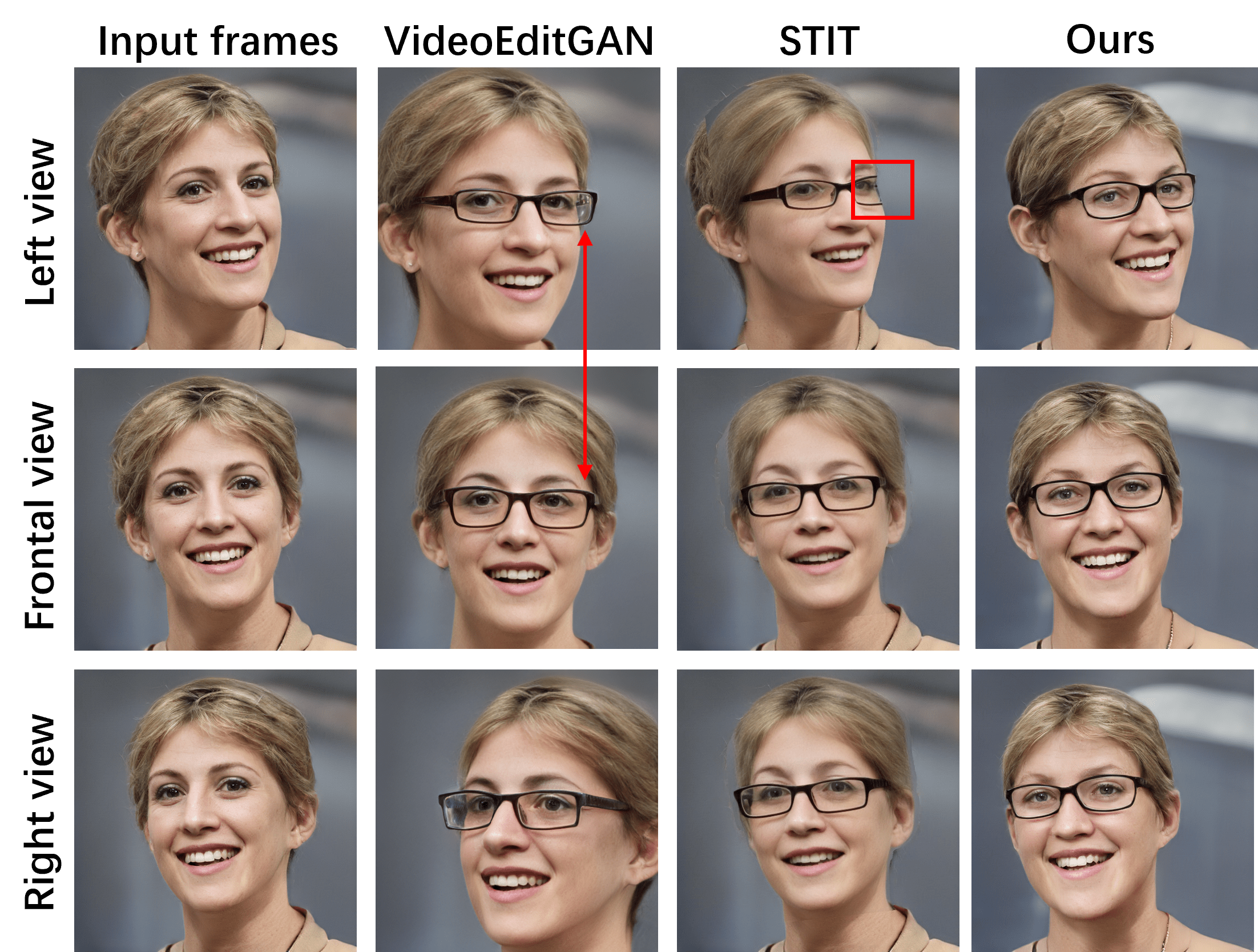}
    \vspace{-0.3in}
    \caption{{\bf Qualitative comparison of 3D consistency.} As indicated by the red box and arrow, it is evident that both STIT and VideoEditGan models are unable to generate 3D view consistency results of comparable quality to ours.}
    \label{fig:comparison2}
    \vspace{-0.2in}
\end{figure}

To examine the 3D consistency of  edited views in an independent manner, a video capturing a static human face from continuously changing camera poses was selected for analysis. In Fig.~\ref{fig:comparison2}, the left, frontal, and right views selected from such video are shown in the left column. After editing the video sequence respectively by VideoEditGAN, STIT, and our method, the 3D reconstruction error from the edited video sequence can be utilized as a metric for assessing the preservation of 3D consistency. 
To perform the multiview stereo reconstruction, we employ COLMAP~\cite{schoenberger2016sfm}. The quantitative outcomes are presented in Table~\ref{tab:comparison_table}. Our method yields the lowest mean reprojection error, a finding that is in line with the visualization results demonstrated in Figure~\ref{fig:comparison2}. All the experiments are done on an RTX3090. 

\begin{table}[h]
\vspace{-0.05in}
\begin{center}
\centering
\caption{{\bf Quantitative comparison on 3D consistency}. 
Mean reprojection error of
the COLMAP reconstruction. 
}
\label{tab:comparison_table} 
\vspace{-0.1in}
\resizebox{0.99\linewidth}{!}{

\begin{tabular}{lccc}
\hline
  & STIT  &VideoEditorGAN &  Ours\\
 \hline
Mean Reproj. Error $ \downarrow$ & 1.1881   & 0.8488  & {\bf 0.7434} \\

\hline
\end{tabular}}
\end{center}
\end{table}

\subsubsection{Temporal Coherence}
\label{sec:Temporal Coherence}
To quantitatively and qualitatively measure the temporal coherence, we select the video sequences from the CelebV-HQ dataset~\cite{zhu2022celebvhq}. Raft~\cite{teed2020raft} is used to estimate the optical flow between two consecutive frames, which serves as a metric for Please refer to the supplementary materials for visualization results and the statistics.

\subsection{Reconstruction}
As a sanity check, it is essential to 
reconstruct the original video from its encoded version. Failure to do so would result in losing the inherent identity of the original video
before any editing can even be performed. Thus, we conducted this experiment on 5 randomly selected videos from the CelebV-HQ dataset~\cite{zhu2022celebvhq}. The comparison is considered to be relatively fair, as none of the methods used in comparison have been trained on the CelebV-HQ dataset. We choose the L2 distance between the feature encoded by VGG16 image encoder~\cite{simonyan2015deep}.
denoted as $\mathcal{L}_v$.
The quantitative comparison is shown in Tab.~\ref{tab:Reconstruction_comparison}.

\begin{table}[h]
\vspace{-0.05in}
\begin{center}
\centering
\caption{{\bf Quantitative comparison of reconstruction}. All the values in the table are multiplied by 100.}
\label{tab:Reconstruction_comparison} 
\vspace{-0.1in}
\resizebox{0.99\linewidth}{!}{

\begin{tabular}{lcccc}
\hline
  & STIT & DVA &VideoEditorGAN &  Ours\\
 \hline
 $\mathcal{L}_v \downarrow$  & 0.47$\pm$0.21  & 0.36$\pm$ 0.07 &0.68$\pm$0.45 &  {\bf  0.31 $\pm$ 0.21 } \\

\hline
\end{tabular}}
\end{center}
\vspace{-0.2in}
\end{table}


\subsection{Ablation Study}
\label{sec:Ablation Study}
We conducted an ablation study on the Latent Encoder Estimator. Specifically, we explored the following alternative setups: (a)~employing a single frame as the input image during the training process, (b)~selecting only one FLAME control to calculate the total loss $\mathcal{L}$ during training, and (c)~our full pipeline implementation, which uses five frames as the input images and five random FLAME controls to calculate the total loss $\mathcal{L}$. In this ablation, we randomly selected 100 subjects from the test dataset. For each subject five random images were used as inputs for setups (b) and (c), while one of the five images is selected as the input for setup (a). Additionally, we selected another image distinct from the five images to calculate the $\mathcal{L}_R$ and $\mathcal{L}_{\mathit{ID}}$. Table~\ref{tab:Latent Encoder ablation} tabulates the resulting statistics of this experiment. As indicated in the table, the full pipeline (c) outperforms other alternative setups and achieves the best results.

\begin{table}[h]
\vspace{-0.1in}
\begin{center}
\centering
\caption{{\bf Ablation study}. The mean value of $\mathcal{L}_R$ and $\mathcal{L}_{\mathit{ID}}$ are multiplied by 10. The standard deviation values of $\mathcal{L}_R$ and $\mathcal{L}_{\mathit{ID}}$ in the table are multiplied by 100.}
\label{tab:Latent Encoder ablation} 
\vspace{-0.1in}
\resizebox{0.80\linewidth}{!}{

\begin{tabular}{lccc}
\hline
  & (a) & (b) & (c)\\
 \hline
 $\mathcal{L}_R \downarrow$  & 1.42$\pm$3.41  & 1.40$\pm$3.04 & {\bf 1.19 $\pm$ 3.10} \\
 $\mathcal{L}_{\mathit{ID}} \downarrow$  & 0.28$\pm$ 1.14  & 0.22$\pm  $0.92 & {\bf 0.16 $\pm$ 0.82}  \\
\hline
\end{tabular}}
\end{center}
\vspace{-0.3in}
\end{table}

\label{sec:formatting}


\section{Conclusion and Discussion}
\label{sec:Conclusion and Discussion}

Our proposed FED-NeRF elevates the face video editing process to operate in a 4D space, thereby ensuring both 3D view consistency and temporal coherence. To the best of our knowledge, this is the first work to tackle the video editing problem by utilizing Dynamic NeRF. Our novel latent Code Estimator utilizes a cross-attention mechanism to aggregate information embedded in multiple frames. The re-engineered Face Geometry Estimator and Stabilizer extract a sequence of facial geometries with good temporal coherence. Working together with our Semantic Editor, our approach presents a significant improvement compared to other video editors which may fall short in preserving geometry consistency across edited frames. We hope our work will inspire further research on solving the 2D video editing problem by incorporating the 4D world representation, which is more aligned with the spatio-temporal reality in which we live.

{
    \small
    \bibliographystyle{ieeenat_fullname}
    \bibliography{main}
}

\clearpage

\setcounter{section}{0}
\renewcommand{\thesection}{\Alph{section}}

\section{Discussion}
\label{sec: Discussion}

\begin{figure*}[t]
    \centering \includegraphics[width=0.90\linewidth]{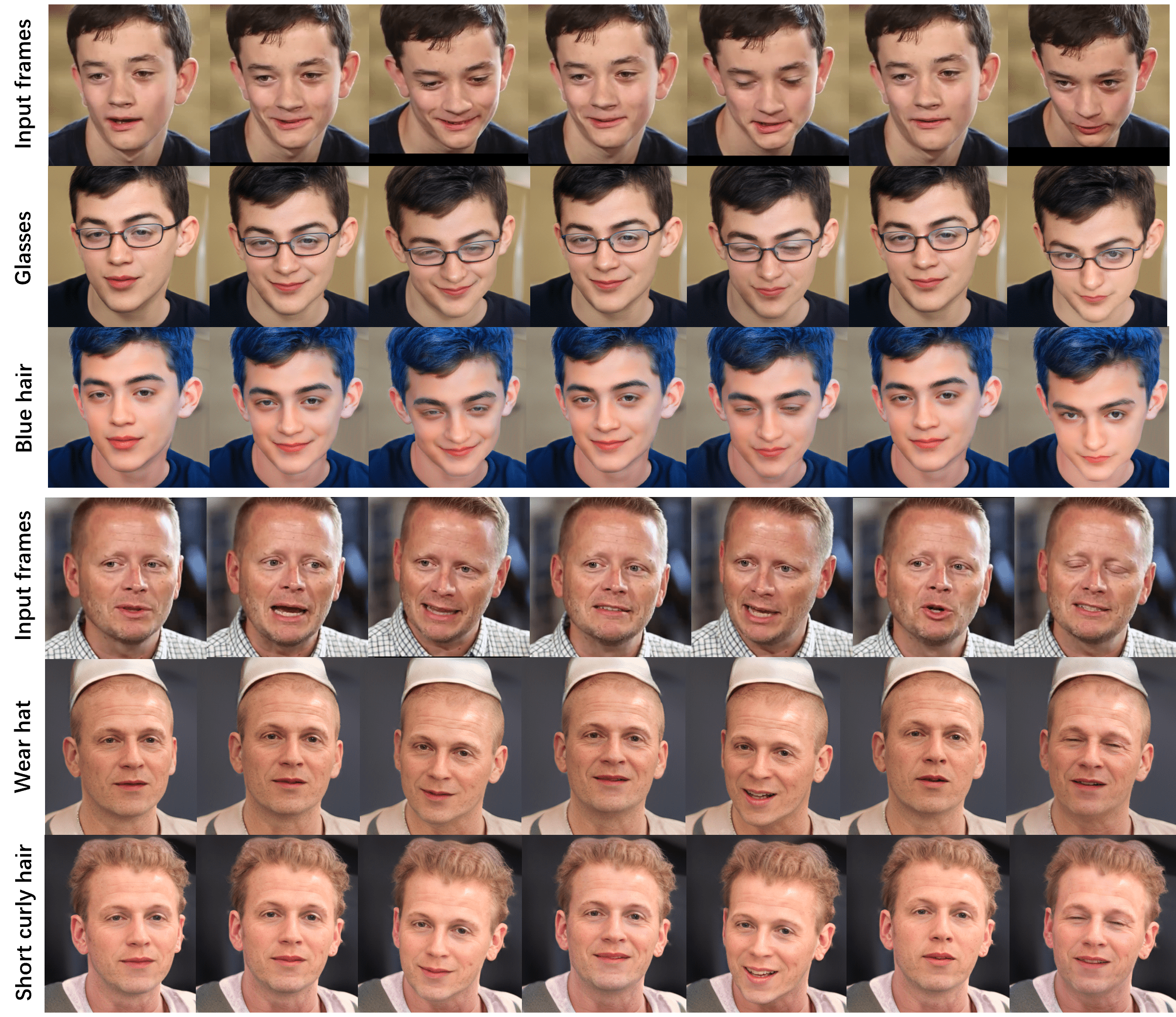}
    \caption{{\bf  More in-the-wild editing results.} These examples show that our model can achieve 3D consistency even when performing
certain edits that alter the facial geometry, such as “Wear a hat”, “Short curly hair”, and so on. }
    \label{fig:more_example_suppl}
\end{figure*}

\begin{figure*}[t]
    \centering
    \includegraphics[width=0.90\linewidth]{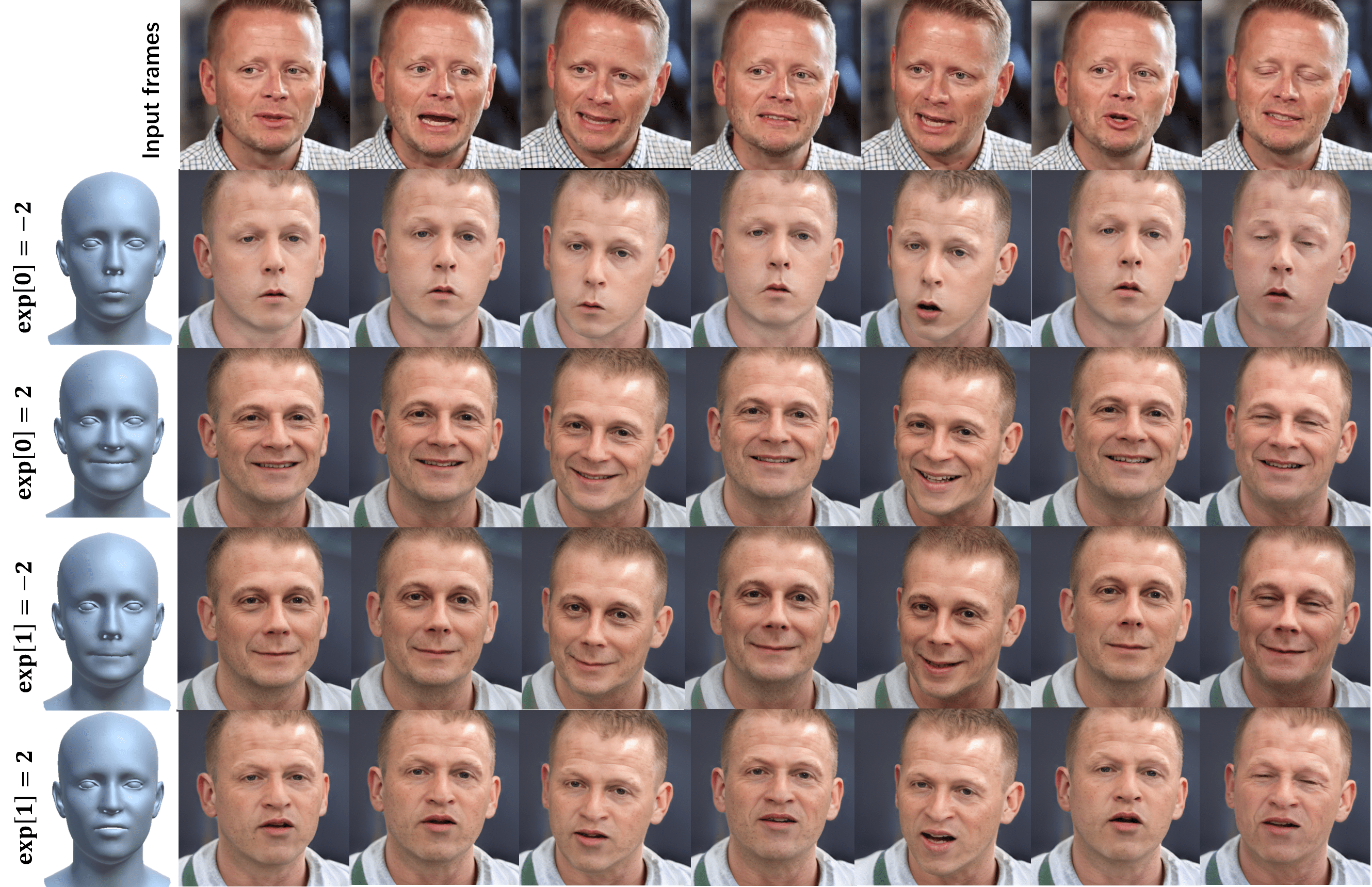}
    \caption{{\bf The Demonstration of the ability to edit the facial expressions explicitly.} The leftmost displays the FLAME mesh, defined by the flame control with a revised value indicated on the left. The other parameters of the flame control maintain consistency with those estimated by our Face Geometry Estimator. }
    \label{fig:explict_expressions}
\end{figure*}

To facilitate the process of face video editing in a 4D space, it is imperative to achieve complete disentanglement between facial geometry and face semantic features. This disentanglement enables seamless semantic editing of the face and explicit control over facial expressions simultaneously. In comparison to other video editing techniques such as ~\cite{tzaban2022stitch,xu2022videoeditgan, Kim_2023_CVPR, yao2021latent}, our approach offers significantly more flexible control over the editing process, albeit at the cost of introducing some distortion to the identity. To the best of our knowledge, our method is the first to tackle the video editing problem using Dynamic NeRF, and it still can achieve comparable semantic editing results while also ensuring 3D consistency in realistic expression editing shown in Fig.~\ref{fig:more_example_suppl}, Fig.~\ref{fig:explict_expressions} and the Demo Video.

\section{More Details on Semantic Editor}
\label{sec: Detailed Information of Semantic Editor}

Analogous to the Latent Mapper architecture presented in StyleCLIP~\cite{Patashnik_2021_ICCV}, distinct layers of the latent code $w^+$  contribute to varying degrees of detail in the generated image~\cite{StyleGAN}. As a result, it is customary to categorize the layers into three distinct groups (coarse, medium, and fine):$ w^+ = (w_c, w_m,w_f)$), and assign each group a specific portion of the (extended) latent vector. The corresponding mapper function can be expressed as follows:
\begin{eqnarray}
    M_t(w^+) = (M_t^c(w_c), M_t^m(w_m),M_t^f(w_t)).
\end{eqnarray}  

The $L_2$ norm of $ M_t(w^+)$ is used to maintain the visual characteristics of the input image. In conjunction with the loss functions $\mathcal{L}_\text{CLIP}$ and $\mathcal{L}_\text{ID}$ introduced in the main paper, the total loss functions can be formulated as follows:

\begin{eqnarray}
    \mathcal{L}(w) = \mathcal{L}_\text{CLIP} + {\lambda}_\text{ID}\mathcal{L}_\text{ID} +  {\lambda}_{L2} \lVert M_t(w^+))\rVert
\end{eqnarray} 
where ${\lambda}_{ID}$ and the ${\lambda}_{L2}$ are the hyperparameters that regulate the strength of  ID preservation and editability, respectively. 

\section{More Details on Temporal Coherence}
\label{sec: Detailed Information of Temporal Coherence}
In order to quantitatively assess the temporal coherence of the proposed method, the Raft algorithm~\cite{teed2020raft} is employed to estimate a dense displacement field between two successive frames $\mathcal{I}_1, \mathcal{I}_2$. The dense displacement field denoted by $(f^1, f^2)$ maps each pixel $(u, v)$ in $\mathcal{I}_1$ to its corresponding coordinates $(u^\prime, v^\prime) = (u+f^1(u), v+f^2(v))$ in $\mathcal{I}_2$. We compute the Euclidean displacement for each pixel $D(u,v) = \sqrt{(f^1(u))^2 + (f^2(v))^2}$. The mean Euclidean displacements among all pixels are denoted as $fl(1,2)$, where the $1$ and $2$ indicate that the mean Euclidean displacement is computed on the first and second frames. To evaluate a video sequence, the initial 40 frames are utilized to compute the mean Euclidean displacements for the sequence, denoted as $flv(1,40)$:

\begin{eqnarray}
    flv(1,40) = \frac{1}{40 - 1} \sum_{i=1,2,..,39} fl(i,i+1)
\end{eqnarray} 

We compared our method with STIT~\cite{tzaban2022stitch} and VideoEditGAN\cite{xu2022videoeditgan} by calculating the metric $flv(1,40)$ on 5 randomly selected video sequences from the CelebV-HQ dataset~\cite{zhu2022celebvhq}. The editing prompt is ``Wear a pair of glasses". The results are shown in Tab. ~\ref{tab:temporal_table}

\begin{table}[h]
\vspace{-0.05in}
\begin{center}
\centering
\caption{{\bf Quantitative comparison on Temporal Coherence}. 
STIT and VideoEditGAN are the abbreviations for ~\cite{tzaban2022stitch} and \cite{xu2022videoeditgan}. 
}
\label{tab:temporal_table} 
\vspace{-0.1in}
\resizebox{0.99\linewidth}{!}{

\begin{tabular}{lccc}
\hline
  & STIT  &VideoEditorGAN &  Ours\\
 \hline
$flv(1,40)$ $ \downarrow$ & 0.5687   & 0.3890  & {\bf 0.3249} \\
\hline
\end{tabular}}
\end{center}
\end{table}

\section{Explicit Editing of Facial Expressions}
\label{sec: Explicitly Edit Facial Expressions}

Our approach provides greater flexibility in video editing compared to other methods~\cite{tzaban2022stitch,xu2022videoeditgan, Kim_2023_CVPR, yao2021latent} due to the clean disentanglement between facial geometry and face semantic features. This allows us to explicitly edit facial expressions in a video by modifying the FLAME controls, whereas other methods are limited to semantic editing. Fig.~\ref{fig:explict_expressions} demonstrates that changing a value of the flame control can directly edit the facial expression.

\section{Implementation Details}
\label{sec: Implementation details}
\begin{figure*}
    \centering
    \includegraphics[width=0.8\linewidth,height=5.5cm]{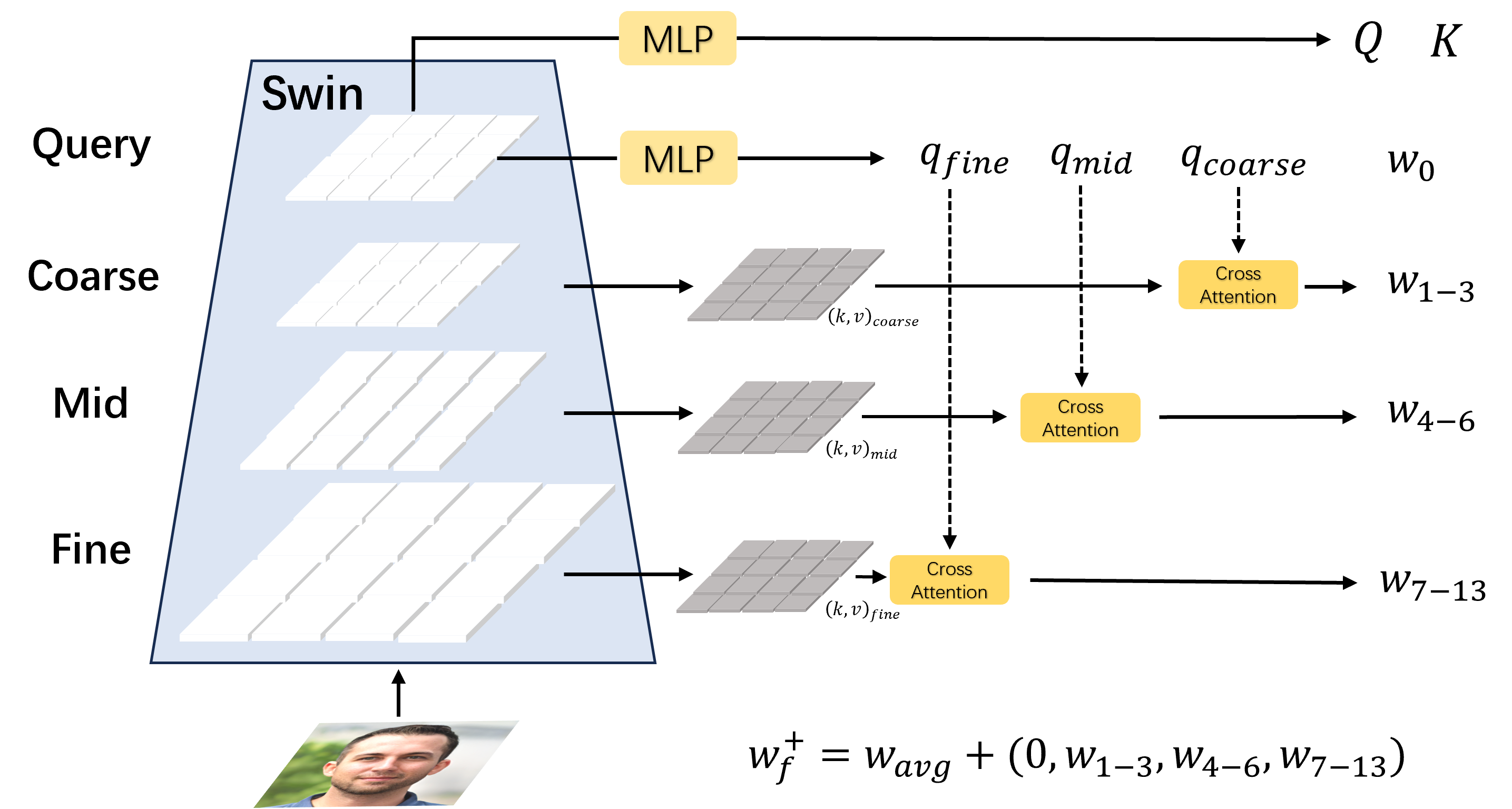}
    \vspace{-0.1in}
    \caption{{\bf The structure of the Image Encoder of Face Geometry Estimator.} }
    \label{fig:image_encoder}
    \vspace{-0.1in}
\end{figure*}

\noindent\textbf{Latent Code Estimator}\quad To balance the overall performance and the GPU memory usage, 5 frames are randomly selected from training datasets and 5 different FLAME controls are used during calculating the loss. The structure of the image encoder is shown in Fig.~\ref{fig:image_encoder}. Inspired by~\cite{yuan2023make},  the intermediate output of the Swin-transformer~\cite{liu2021Swin} is split into four levels “query, coarse, mid, and fine”. “query” is used to get $w0$, $q_\text{fine}$, $q_\text{mid}$, and $q_\text{coarse}$. “coarse”, “mid”, “fine” layers are used to obtain keys and values $(k,v)_\text{coarse}$, $(k,v)_\text{mid}$, and $(k,v)_\text{fine}$. Then these level queries with their corresponding keys and values are sent into cross-attention layers to produce different $w_i$.
The $Q, K$ are extracted from the “query" layer by MLP layers, since the $Q, K$ contains the information on how to merge multiple $w^{+\prime}$'s, which is high-level information and thus should be extracted from the latter layers of the pyramid features. We trained the Latent Code Estimator on 2 V100 GPUs for nearly 2 weeks with batch size $2$. The Adam optimizer~\cite{kingma2017adam} is used with an initial learning rate of $7 \times 10^{-5}$.

\noindent\textbf{Face Geometry Estimator}\quad The learning rate is $5 \times 10^{-5}$ which is decayed by for each epoch. We trained the Face Geometry Estimator on 4 RTX3090s for 2 days with batch size $4$.

\noindent\textbf{Semantic Editor}\quad We trained the mapper with the following settings: ${\lambda}_{L2} = 0.7$,  ${\lambda}_\text{ID} = 0.1$, maximum steps $=50000$, batch size $=4$, and learning rate $ \in [1.0,1.5,2.0,2.5,3.0,3.5] $. The learning rate depends on on the editing test prompt. A large change in facial expressions usually needs a larger learning rate. 

\section{Demo Video}
\label{sec: video}
As the focus of our method is  video editing, a demonstration video is  a more effective means for evaluating the performance. Our demo video accompanying the supplementary material contains 1) comparisons with other methods, 2) explicit facial expression editing, and 3) in-the-wild video editing. 


\end{document}